\documentclass[journal,twoside,web]{ieeecolor}
\usepackage{generic}
\usepackage{cite}
\usepackage{amsmath,amssymb,amsfonts}
\usepackage{algorithmic}
\usepackage{graphicx}
\usepackage{algorithm,algorithmic}
\usepackage{hyperref}
\hypersetup{hidelinks}
\usepackage{textcomp}
\usepackage{bm}
\usepackage{multirow}
\usepackage{xurl}
\def\BibTeX{{\rm B\kern-.05em{\sc i\kern-.025em b}\kern-.08em T\kern-.1667em\lower.7ex\hbox{E}\kern-.125emX}}
\begin{document}
\title{CorVS+: Correspondence-Driven Association of Video Trajectories and Sensors for Identity- Aware Person Localization in Warehouses}
\author{
  Kazuma Kano,
  Yuki Mori,
  Shin Katayama,
  Kenta Urano, \IEEEmembership{Member, IEEE}, \\
  Takuro Yonezawa, \IEEEmembership{Member, IEEE},
  and Nobuo Kawaguchi, \IEEEmembership{Member, IEEE}
  \thanks{
    % Received 2 May 2018; revised 9 September 2018; accepted 12 October 2018.
    % Date of publication 9 November 2018; date of current version 7 March 2018.
    This work is supported in part by JSPS KAKENHI (JP22K18422), NEDO (JPNP23003), CSTI SIP3 (JPJ012495), and JST BOOST (JPMJBS2422).
    This paper was presented in part at the Fifteenth International Conference on Indoor Positioning and Indoor Navigation, Tampere Hall, Tampere, Finland, September 2025.
    \textit{(Corresponding author: Kazuma Kano.)}
  }
  \thanks{
    This work involved human subjects in its research.
    The authors confirm that all human subject research procedures and protocols are exempt from review board approval.
  }
  \thanks{
    The authors are with the Graduate School of Engineering, Nagoya University, Nagoya, Aichi 464-8603, Japan (e-mail: \href{mailto:kazuma@ucl.nuee.nagoya-u.ac.jp}{kazuma@ucl.nuee.nagoya-u.ac.jp}).
  }
  \thanks{
    Data and code will be available on-line at \url{https://doi.org/10.5281/zenodo.17745683} and \url{https://github.com/kazumakano/corvs-plus}.
  }
}

\maketitle

\bstctlcite{no-dash}

\begin{abstract}
  Logistics warehouses have struggled with labor shortages, but the inbound processes remain particularly human-powered.
Worker location data is a key to higher productivity in such cases.
Fixed cameras are a promising tool for localization, as they also offer valuable environmental information such as package status.
However, identifying individuals from visual data alone is often impractical.
To enable identity-aware localization, prior studies have attempted to identify people in videos by associating their trajectories with wearable sensor measurements.
Although this appearance-independent approach has several advantages, existing methods may fail under real-world conditions.
Therefore, we propose CorVS+, a novel data-driven person identification framework based on the correspondence between visual tracking trajectories and sensor measurements.
Firstly, our deep learning model predicts the correspondence probabilities and reliabilities for every pair of a trajectory and sensor measurements.
Secondly, our algorithm matches the pairs over time based on the model predictions.
We developed a dataset comprising 27 hours of sensor measurements and 38 km of trajectories in a warehouse.
This dataset covers actual activities and challenging situations, such as multiple stationary workers inspecting items.
The evaluation indicated the superiority of CorVS+ over existing methods and the effectiveness of its unique designs for industrial-scale settings.
The model and dataset will be available at \url{https://doi.org/10.5281/zenodo.17745683}.

\end{abstract}


\section{Introduction}
Digital transformation in industrial sites has attracted attention, driven by demand for higher productivity and work quality \cite{11058642,10564577}.
Logistics warehouses are among the workplaces most affected by labor shortages due to the expansion of e-commerce markets and consequent increase in workload.
However, full automation with robots is not feasible, given the need to handle heterogeneous item sizes and shapes, as well as to accommodate fluid market volumes and trends.
Thus, many warehouses still rely heavily on human workers, particularly within complex inbound processes.
In this context, worker location data is crucial for improving work visibility and efficiency.
The data offers potential for various applications beyond navigation, such as shift planning \cite{10.1287/trsc.2020.1029,11342838}, dynamic task assignment \cite{Calzavara2024,YU2025104082}, and layout optimization through simulation \cite{Aslan16022025,11342838}.

\begin{figure}[tb]
  \includegraphics[width=\linewidth]{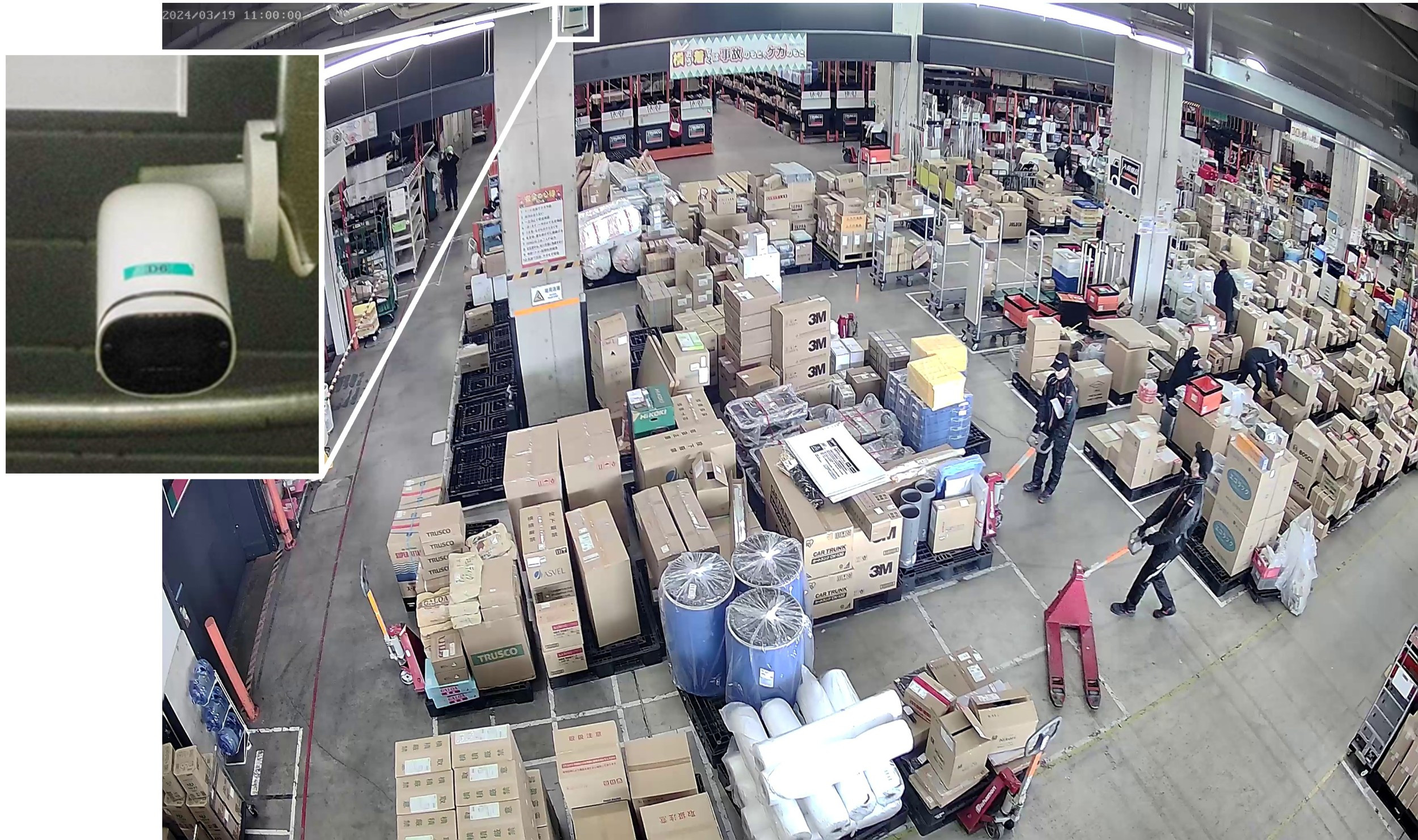}
  \caption{Camera mounted on warehouse ceiling.}
  \label{fig:env}
\end{figure}

\begin{figure*}[tb]
  \centerline{\includegraphics[width=\linewidth]{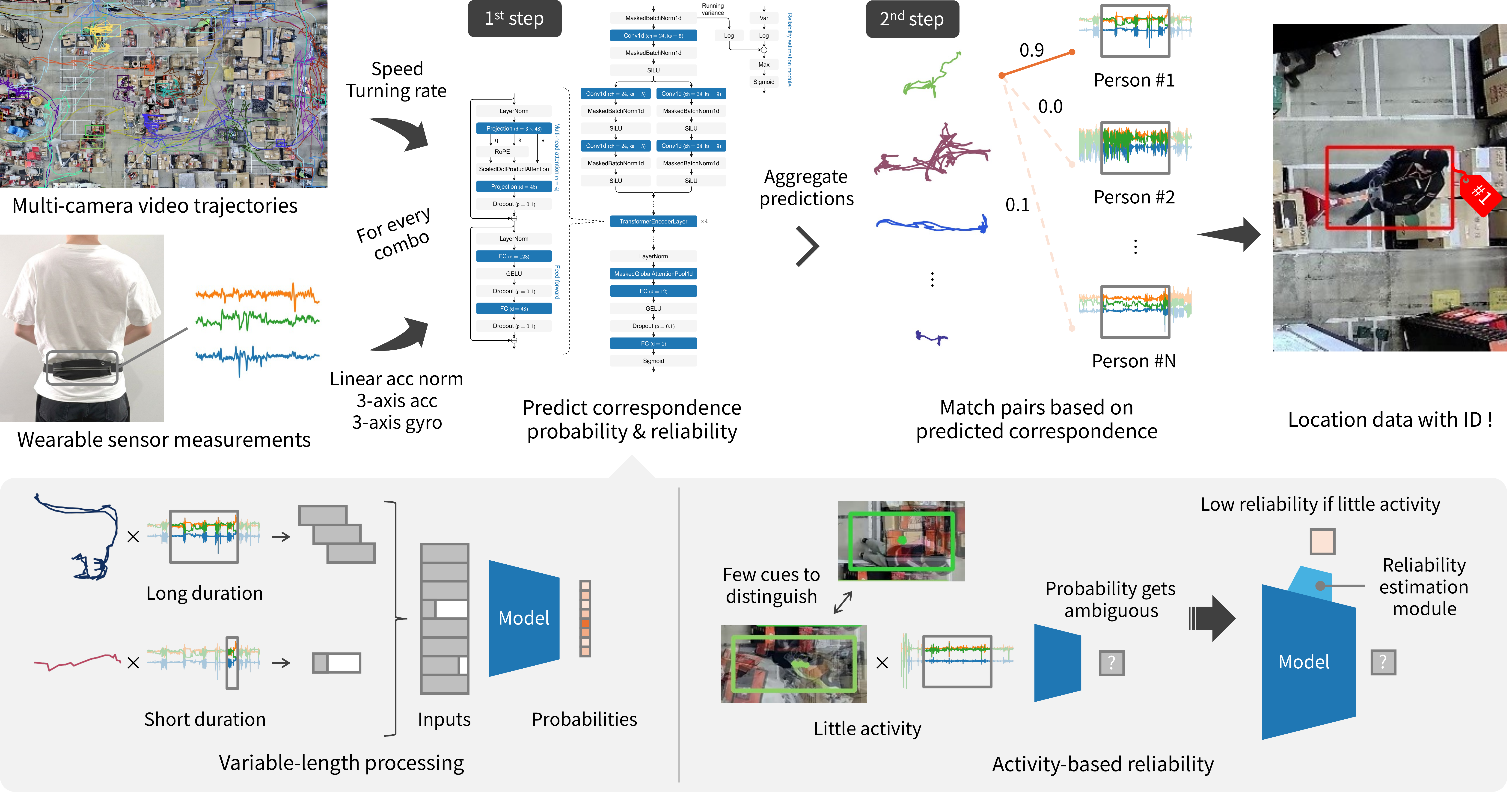}}
  \caption{Identification process of CorVS+.}
  \label{fig:overview}
\end{figure*}

A wide range of approaches for indoor localization has emerged over the years \cite{10561484}.
Trilateration based on Received Signal Strength (RSS) of BLE or Wi-Fi is relatively inexpensive and widely used in prior studies \cite{10.1007/978-3-031-05463-1_21,10332219,10332522}.
Nevertheless, dense clutter and obstacles, including packages and human bodies, induce multipath fading and shadowing frequently in warehouses.
The trilateration tends to be less accurate in such environments.
Fingerprinting is another common approach based on radio frequency signals \cite{NGUYEN2024101912,10.1145/3712277,11213392}.
However, it is also unsuitable for warehouses because the frequent turnover of packages constantly alters the radio maps.
% While various other approaches, such as Angle of Arrival (AoA) \cite{10364949,10786160}, Angle of Departure (AoD) \cite{10364949}, Round Trip Time (RTT) \cite{10364949,10.1145/3748777.3748809}, and Phase-Based Ranging (PBR) \cite{10786118,10755009}, have been developed, they typically require transceivers with specific hardware capabilities.
While various other approaches, such as Angle of Arrival (AoA) \cite{10364949,10786160}, Round Trip Time (RTT) \cite{10364949,10.1145/3748777.3748809}, and Phase-Based Ranging (PBR) \cite{10786118,10755009}, have been developed, they typically require transceivers with specific hardware capabilities.

% \begin{figure}[tb]
%   \includegraphics[width=\linewidth]{figures/fingerprint}
%   \caption{Received signal strength from BLE beacon.}
%   \label{fig:fingerprint}
% \end{figure}

In this study, we employ cameras mounted on a ceiling, as shown in Fig. \ref{fig:env}.
Cameras are advantageous in providing not only absolute human locations but also contextual information, such as the status of packages and equipment \cite{10412228,10494498}.
Nevertheless, identifying individuals with only visual data is often impractical.
It necessitates integration with other modalities for advanced identity-aware applications.
Accordingly, some prior studies identified people in videos by comparing visual tracking trajectories with wearable sensor measurements \cite{10.1145/1579114.1579134,10.1145/3302509.3311057,10.1145/3395035.3425968,10.1145/2663204.2663270,5289412,9166762,10.1145/3579832,7266712,10.1145/2733373.2806301,8373408,8935339,s24113680}.
Smartphones can be a particularly cost-effective solution here because they function as handy terminals for logistics operations in addition to sensors for localization and task recognition \cite{MASTAKOURIS2023115,10.1145/3714394.3756198}.
In fact, handy terminals running Android OS have already entered the market.
However, the existing identification methods may break down in real-world settings due to restrictive scenario assumptions, insufficient robustness to complex motions, etc.

To address these challenges, we designed a novel data-driven method, CorVS+, grounded in on-site studies.
It identifies visually tracked subjects wearing sensors through two steps, as illustrated in Fig. \ref{fig:overview}.
First, it predicts correspondence probabilities and their reliabilities by deep learning for every pair of a trajectory and simultaneous sensor measurements.
Second, it matches the trajectories and sensor measurements based on the predicted probabilities and reliabilities.
We developed a dataset comprising trajectories and sensor measurements of actual warehouse workers.
And then, we demonstrated the superiority of CorVS+ over existing methods.
This paper is an extended version of our preliminary work \cite{11213222}.
While building upon its core framework, this paper introduces variable-length modeling and data augmentation, expands the dataset, and provides more comprehensive evaluations.
Furthermore, the pre-trained model and dataset will be released at \url{https://doi.org/10.5281/zenodo.17745683} to support reproducibility and also foster research toward real-world applications.
Our contributions are summarized below.
\begin{itemize}
  \item We propose a deep learning model and training strategies for estimating correspondences directly from trajectory features and sensor measurements.
        This model is unique in several respects, including its variable-length processing and consideration of subjects' activeness.
  \item We propose a matching algorithm that incrementally associates the pairs based on the estimated probabilities and reliabilities, addressing practical situations.
  \item We created and will open up an unprecedented practical dataset comprising 27 hours of sensor measurements and 38 km of trajectories collected in a warehouse.
  \item We present evaluation metrics for person identification with a high presence of non-target individuals.
        The evaluation validated the method and derived empirical insights.
\end{itemize}

\section{Related Work}
% \subsection{Person Re-identification with Fixed Cameras}
% Person Re-Identification (Re-ID) is a key task in computer vision that matches the same individuals across different views based on their appearances \cite{10858128}.
% Mainstream studies use representation learning to embed the visual features \cite{9011001,9710179}.
% In addition, there is growing interest in pre-training with unlabeled large-scale datasets \cite{10378315,10543111}.
% Recent studies also leveraged semantic knowledge of Vision--Language Models (VLM) and demonstrated high accuracy \cite{10.1609/aaai.v37i1.25225,3737916.3739368}.
% Nevertheless, note that Re-ID does not necessarily determine the exact identities.
% Although Re-ID is instrumental for tracking \cite{10208715} or anonymous analysis, it is insufficient for more detailed examinations.

\subsection{Person Identification with Fixed Cameras}
A simple means to find specific individuals in videos is to get them to wear markers like AprilTag \cite{5979561}.
It can differentiate individuals regardless of their appearances as long as there are enough patterns.
However, marker recognition assumes adequate image quality and marker orientation.
On the other hand, various studies employed visual attributes for identification, such as faces \cite{10522363,10550878}, body types \cite{10744482}, and other soft biometrics \cite{attr-based-people-search}.
These approaches have potential applications in security and investigation but also require high resolution and appropriate angles.
In particular, overhead angles make identification difficult due to the lack of visual features.
Gait recognition, which does not rely on such high resolution, has also been explored for identification \cite{10735362}.
Nevertheless, task-specific movements often overshadow individual gait traits in industrial settings.
This circumstance makes gait recognition ineffective.

\subsection{Person Identification with Fixed Cameras and Wearable Sensors}
Since identification solely based on appearance brings impractical constraints, prior studies sometimes incorporated wearable sensors.
It enables the tagging of people in videos by matching visual tracking trajectories with corresponding sensor measurements.
This approach works even when people wear uniforms and exhibit limited visual variations.
Moreover, it raises fewer privacy concerns than appearance-based approaches like face recognition, as it does not require profile databases for identity verification.
Teixeira et al. identified individuals based on Pearson correlation coefficients between horizontal acceleration magnitude calculated from trajectories and measured by sensors \cite{10.1145/1579114.1579134}.
Akbari et al. \cite{10.1145/3302509.3311057} and Ishihara et al. \cite{10.1145/3395035.3425968} also compared acceleration magnitude from cameras and sensors.
However, the second-order differentiation at converting trajectories to acceleration may amplify the errors and reduce identification accuracy, especially in noisy conditions.

On the other hand, Henschel et al. employed cosine similarity of horizontal body orientations estimated from videos and sensor measurements in addition to acceleration \cite{9166762}.
The shapes of visual tracking trajectories do not matter in this approach, as the orientations in videos can be recognized independently per frame.
However, it requires visual recognition models for orientation estimation as well as person detection.
Moreover, it collapses unless the sensors are fixed to the bodies.
In environments like warehouses, where steel structures and metal equipment are prevalent, magnetic anomalies can also compromise the sensor orientation estimation considerably.
Bannis et al. proposed IDIoT, which uses sensor rotations from sensor measurements and body pose changes from videos to identify who and which parts are equipped with the sensors \cite{10.1145/3579832}.
This method is agnostic to whether the sensors are attached or held.
However, it still needs pose estimation tools such as OpenPose \cite{8765346}, which often suffer from top-down views.

Several studies applied Pedestrian Dead Reckoning (PDR) techniques to identification.
Jiang et al. computed similarity transformation matrices that align visual tracking trajectories with PDR trajectories and associated similar pairs \cite{7266712}.
Nagai et al. discriminated between people carrying sensors and others by comparing the counts of walking events detected in visual tracking and PDR \cite{10.1145/2733373.2806301}.
Zhang et al. compared steps and headings estimated from videos and sensor measurements \cite{8373408}.
Li et al. proposed iPAC, which matches trajectories from visual tracking and PDR based on walking events and headings \cite{8935339}.
However, one of the common challenges among these studies is the lack of robustness to complex motions.
The conventional PDR methods used in these studies struggle to handle actions such as squatting or backward walking, which frequently occur in warehouses.
While PDR methods based on deep learning deliver improved robustness \cite{Rao_Kazemi_Ding_Shila_Tucker_Wang_2022,gait-robust-heading-estim,Nguyen_Le_Havinga_2025}, precise location labels for model training are hard to obtain in industrial settings.

Another intuitive approach based on deep learning is to design models that receive visual tracking trajectories and sensor measurements and estimate their correspondences in an end-to-end manner.
% The task simplification from regression in PDR to binary classification in this approach facilitates training with noisy trajectory data.
This approach facilitates training with noisy trajectory data through the task simplication from regression in PDR to binary classification.
Yan et al. identified individuals based on correspondence probabilities predicted by a deep learning model \cite{s24113680}.
However, there are still substantial gaps when applying this method to real-world scenarios.
For example, it does not consider situations where multiple people are stationary with few clues to distinguish.

\section{Proposed Method: CorVS+}
\subsection{System Overview}
In this study, we propose CorVS+, a data-driven method that identifies people in videos with wearable sensors via Correspondence between the Visual tracking trajectories and Sensor measurements.
It provides absolute location information linked to the identities and is feasible in practical environments like warehouses.
The identification process consists of two stages: correspondence estimation and matching.
First, it estimates correspondence probabilities and their reliabilities with a deep learning model for every pair of a visual tracking trajectory and simultaneous sensor measurements.
Second, it matches the trajectories and sensor measurements based on the estimated probabilities and reliabilities.
CorVS+ does not rely on appearance and is compatible with arbitrary tracking systems, including edge AI cameras \cite{10628356,10976142}.

\subsection{Correspondence Estimation Model}
Recent progress in computer vision technologies has improved the performance of human detection and tracking \cite{NEURIPS2024_c34ddd05,10943837,10.1609/aaai.v38i6.28386}.
Still, location data calculated from the bounding boxes often contain errors, particularly with distorted wide-angle cameras or occlusion-prone environments.
Additionally, it is too costly to manually implement a heuristic rule set that can handle a wide range of personal attributes and actions in industrial settings.
In this context, we leverage deep learning to consider various spatiotemporal features automatically and improve the robustness.
Key extensions regarding the model from our preliminary work \cite{11213222} include the following.
\begin{itemize}
  \item The new model accepts variable-length inputs. This change involves architectural updates, such as padding support and positional encoding.
  \item Two data augmentation strategies, random masking and shifting, are applied to positive samples.
  \item The pre-trained weights will be made publicly available.
\end{itemize}

\subsubsection{Input and Output}
The input modalities are listed below.
We adopt movement speeds and linear acceleration magnitude, which the prior studies commonly used as identification cues.
These data reflect movement intensity well; linear acceleration indicates movements themselves by excluding gravity effects.
We also employ turning rates, acceleration, and angular velocity as the inputs.
It intends to provide information regarding movement headings, sensor orientations, and other key factors.
\begin{enumerate}
  \renewcommand{\theenumi}{\alph{enumi}}
  \item Movement speeds calculated from visual tracking
  \item Turning rates calculatd from visual tracking
  \item Linear acceleration norm measured by inertial sensors
  \item 3-axis acceleration measured by inertial sensors
  \item 3-axis angular velocity measured by inertial sensors
\end{enumerate}
As preprocessing, we smooth these data by applying the Gaussian filter with a standard deviation of 0.2 seconds and resample them at 10 Hz.
We empirically selected these values to balance information retention and noise reduction, helping the model focus on meaningful features.
In general, cameras have lower sampling rates than inertial sensors.
We align their time scales by upsampling the trajectory features and downsampling the sensor measurements.

\begin{figure}[tb]
  \centerline{\includegraphics[width=\linewidth]{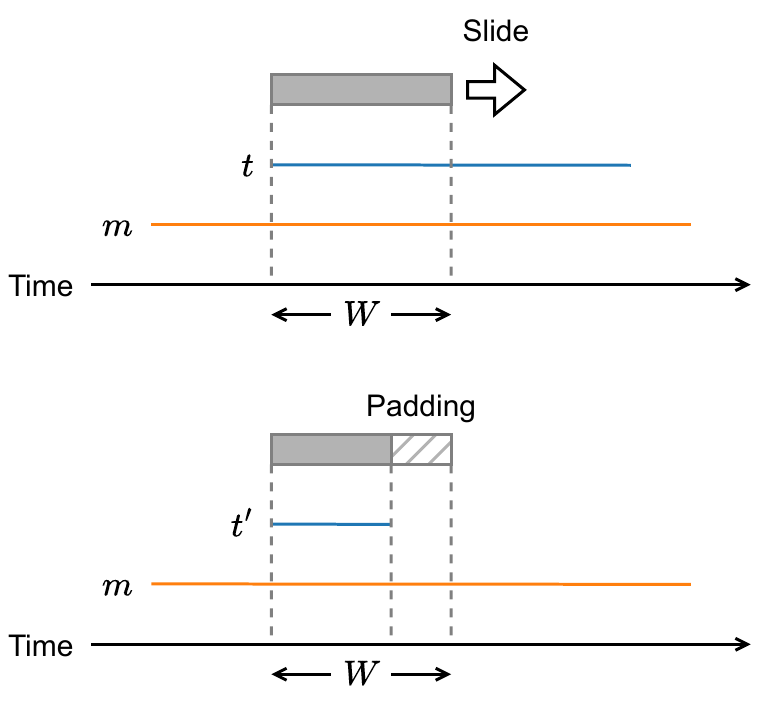}}
  \caption{Sliding window and padding processes.}
  \label{fig:win}
\end{figure}

Then, a sliding window of length $W$ retrieves data segments from temporally overlapping regions of the trajectory feature and sensor measurement sequences, as schematized at the top of Fig. \ref{fig:win}.
The blue and orange lines denote spans with a trajectory $t$ and sensor measurements $m$ present.
The gray box represents a range for the model input sampled by the window.
While a longer input enriches information and enhances the model performance, it constraints applicability to fragmented short trajectories.
To reconcile such a trade-off, we adopt variable-length inputs.
Samples shorter than the window length $W$ are padded to maintain a uniform length, as depicted at the bottom of Fig. \ref{fig:win}.
This formatting allows for parallel computation via batching.
In this paper, we set the $W$ to 600 (i.e., 1 minute).
The window stride is 10 for training and 1 for validation.
% Additionally, filtering out very short samples, which include insignificant data from people appearing briefly at frame edges, tended to yield better results.
% To this end, we impose a minimum input length of 100 (i.e, 10 seconds).
Note that, since some input modalities, such as acceleration, are quantitative variables that can take arbitrary values, padding with a fixed value is insufficient to distinguish the padded regions from them.
We thus pass binary masks indicating the valid regions to the model as well as the inputs.
In the prediction phase, we feed these data into the model for every combination of a visual tracking trajectory and simultaneous sensor measurements.

The outputs are two scalars: probability and reliability.
The probability denotes how likely the tracked subject corresponds to the sensor wearer.
However, this correspondence gets inherently ambiguous when there is little activity in both the trajectory and sensor signals.
For example, given a trajectory and sensor signals of two stationary individuals, the model may incorrectly infer that they are identical.
In fact, many workers in warehouses remain in fixed locations for long durations to inspect items.
To address this challenge, we introduce activity-based reliability of the estimated probability, separate from the internal confidence.
The subsequent process uses these scores to match the pairs.

\subsubsection{Architecture}
In the contexts of Human Activity Recognition (HAR) and PDR using wearable sensors, most scenarios assume that sensor measurements are constantly available, and studies into deep learning models with variable input length remain limited \cite{10.1145/3699770,10746455,10.1145/3706598.3713721}.
Meanwhile, inference from variable-length sequences is well-established in domains such as natural language processing and speech processing.
We draw on those techniques to implement a model for our targeting task, correspondence estimation between visual tracking trajectories and sensor measurements.

There are several paradigms for fusing different modalities \cite{10123038}.
One well-known paradigm uses independent encoders for each modality to map the data into a common representation space, exemplified by CLIP \cite{pmlr-v139-radford21a} and ImageBind \cite{10203733}.
Identity verification becomes theoretically possible by comparing the embedded vectors.
Nevertheless, this approach is considered inappropriate for visual tracking trajectories and sensor measurements.
The information in these data is so asymmetric that the projection into a single space would discard much of it before the identity verification.
In addition, this approach does not explicitly consider temporal correspondence between the modalities.
With these points in mind, we adopt an early fusion paradigm that synchronizes and merges the data at the outset to focus on interactions across the modalities.

\begin{figure}[tb]
  \centerline{\includegraphics[width=\linewidth]{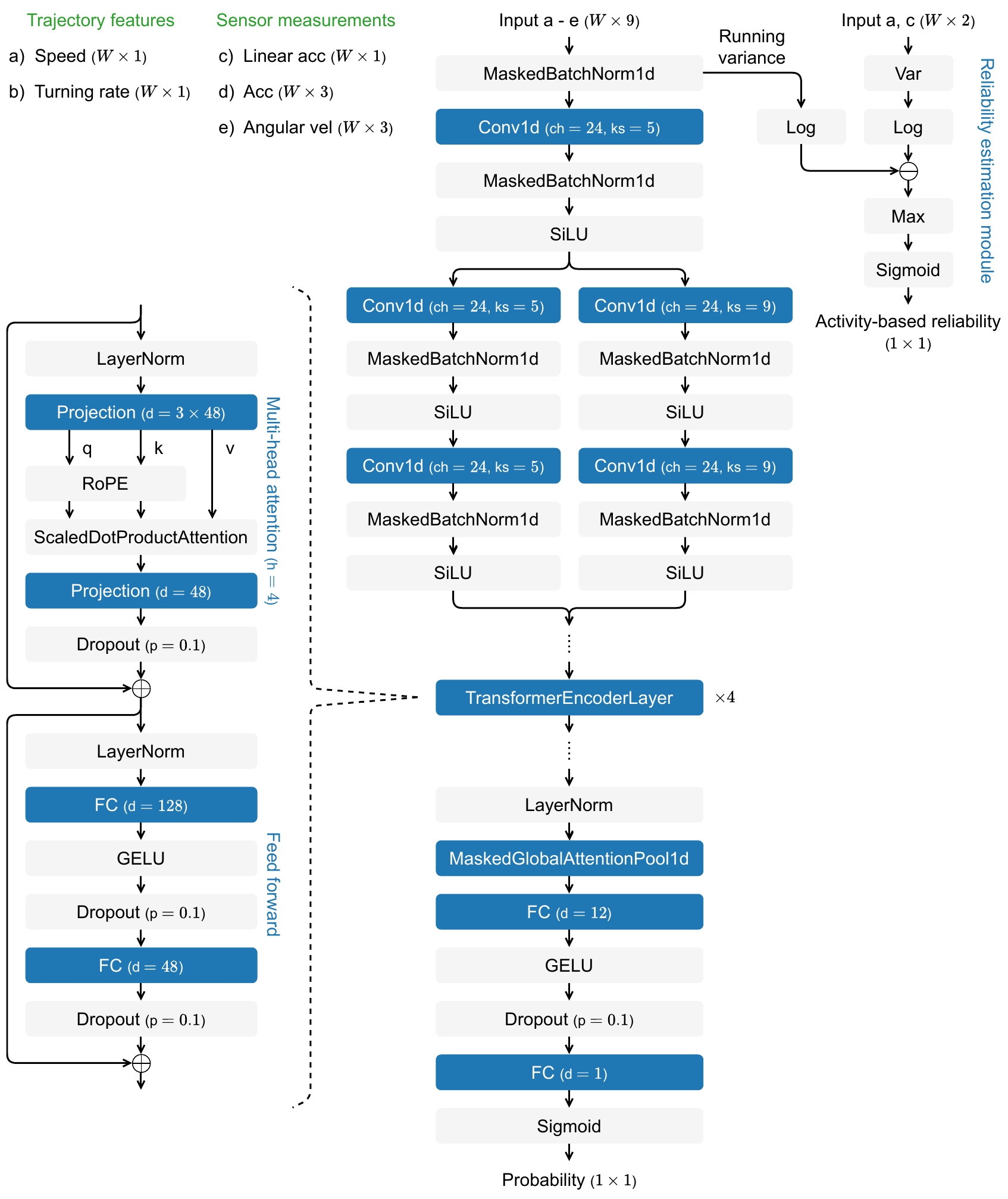}}
  \caption{Correspondence estimation model architecture.}
  \label{fig:model}
\end{figure}

Fig. \ref{fig:model} depicts the architecture with tuned hyperparameters.
We extend DualCNN-Transformer from our previous work on PDR \cite{gait-robust-heading-estim}.
It can capture multi-timescale features through the two different-sized convolutional paths and self-attention blocks.
It helps recognize both short-term actions like squatting and long-term movements like walking.
The initial batch normalization layer serves as online data standardization and mitigates scale divergence across the modalities.
The subsequent batch normalization layers aim to stabilize the learning process and also improve the generalization performance \cite{10.5555/3327144.3327174}, as with general CNNs.
We omit the affine parameters only for the initial one to avoid functional redundancy.
Note that all these batch normalization layers are customized to compute means and variances exclusively from valid values, ignoring the padded regions; we refer to this as masked batch normalization.

We incorporate Rotary Position Embedding (RoPE) \cite{10.1016/j.neucom.2023.127063}, a representative scheme for injecting relative positional dependencies into attention mechanisms.
RoPE is also a standard in modern Large Language Models (LLMs) and Vision--Language Models (VLMs), such as Gemma \cite{gemmateam2025gemma3technicalreport} and Llama \cite{grattafiori2024llama3herdmodels}.
This choice reflects the intuition that the correspondence probability relies on mutual relationships among features rather than their exact timing.
For example, the correspondence estimation should be invariant to whether an informative feature pattern appears in the former or latter part of the time window.
Additionally, RoPE does not require learnable parameters for every absolute position.
It enables seamless adaptation to arbitrary sequence lengths, even those not seen during the training.
This scalability is advantageous for handling variable-length inputs.
Note that all other learnable parameters are shared across absolute positions as well.
It means the model size depends not on its temporal length but only on its dimensional width.

After the CNN backbone and Transformer encoder, we apply global attention pooling to aggregate the features along the time dimension.
Features at each time often do not contribute equally to correspondence estimation.
Some intervals can be uninformative or misleading due to noise, whereas others may provide discriminative cues.
In this case, average pooling, which uniformly dilutes the features, is suboptimal.
Attention pooling allows the model to dynamically assign importance to each time and flexibly fuse features relevant to the correspondence estimation.
This layer is implemented as a single-head architecture and also supports padded data; we refer to this as masked global attention pooling.
The attention layers of the Transformer encoder and the masked global attention pooling layer overwrite their attention scores for the padded regions with $-\infty$ \cite{10.5555/3295222.3295349}.
It prevents padding values from contributing to the final outputs and affecting the weight updates through backpropagation.
This practice is common for processing padded sequences with Transformers.

Another key novelty is attaching a non-parametric module for the reliability estimation.
It receives the segments of movement speeds $\bm{x}_{spd}$ and linear acceleration norm $\bm{x}_{acc}$ from the original inputs and their running variance $\tilde{\sigma}_{spd}^2$ and $\tilde{\sigma}_{acc}^2$ from the initial batch normalization layer.
Then, it computes logarithms of the input variance over the running variance for the speeds and linear acceleration each.
At last, it yields the activity-based reliability $r$ as a sigmoid of the greater one.
\begin{equation}
  r = \operatorname{sigmoid}\left(\max\left(\log\frac{\operatorname{var}\left(\bm{x}_{spd}\right)}{\tilde{\sigma}_{spd}^2}, \log\frac{\operatorname{var}\left(\bm{x}_{acc}\right)}{\tilde{\sigma}_{acc}^2}\right)\right)
\end{equation}
We interpret the variance of speeds and linear acceleration as activity levels of the visual tracking trajectory and sensor signals, respectively.
The reliability implies whether at least one of the trajectory and sensor signals exhibits much activity compared to typical.
More precisely, it takes 0 if both exhibit minimal and approaches 1 if either exhibits more.
This formulation offers an objective quantification of reliability, regardless of how often stationary pairs of trajectories and sensor measurements actually belong to the same individuals in the training data.

\subsubsection{Training}
The training needs positive and negative pairs of trajectories and sensor measurements.
We construct the negative pairs by randomly coupling data from different individuals or timestamps.
Here, we exclude the trajectories of people without sensors during training.
This curation aims to emphasize learning inter-modal relationships rather than per-modal patterns.
In addition, we cap the ratio $\rho_{neg}$ of negative to positive samples to avoid combinatorial explosion and stabilize the learning process.
The negative sample ratio $\rho_{neg}$ is set to a large number for training while fixed to 1 for validation.

Because the positive samples are much scarcer relative to the negatives, we apply two data augmentation strategies to the positives.
First, we randomly mask parts of the inputs, forcing the model to estimate from incomplete data.
Because attention mechanisms upweight salient regions, the model tends to be optimized for them preferentially.
This temporal masking aims to explore also subtle feature patterns and encourage multi-faceted inference, analogous to dropout.
Here, adjacent time points are likely to correlate, so we mask a contiguous block rather than scattered points to ensure the regularization effect.
In this paper, we determine the masking proportion such that the combined length of the masked or padded region accounts for 25\% of the maximum input length $W$.
We do not perform the masking for samples shorter than 75\% of the $W$.

Second, we randomly shift only sensor measurements along the time dimension.
In practical settings, time synchronization between cameras and sensors is seldom perfect.
The timestamp misalignment typically consists of two components: a systematic offset inherent to the device type or measurement protocol, and a stochastic jitter dependent on individual device variations or timing.
The model is expected to implicitly learn to compensate for the systematic offsets through the training.
Accordingly, this temporal shifting intends to develop invariance specifically to the stochastic jitter.
In this paper, the shift length follows a normal distribution with a mean of 0 seconds and a standard deviation of 0.25 seconds.
We empirically set this value based on a preliminary investigation into the clock drift and time synchronization behavior across 10 smartphones we employ as sensors.

Using these data, we train the model to output the probability 1 if identical and 0 otherwise.
The reliability estimation module is not involved in the training.
% In this paper, we employ Binary Cross Entropy (BCE) as a loss function.
% We apply weights to loss values for the positive samples according to the proportion to balance the contribution of the positive and negative samples.
We employ Binary Cross Entropy (BCE) as a loss function and apply weights to loss values for the positive samples according to the proportion.
We also experimented with Focal loss \cite{8417976}, often used in class-imbalanced tasks such as object detection, but it did not result satisfactorily.
The model may have focused on low-activity data and did not learn effectively.
Separately, we use AdamW \cite{loshchilov2018decoupled} as an optimizer and cosine annealing \cite{loshchilov2017sgdr} with linear warmup \cite{goyal2018accuratelargeminibatchsgd} as a learning rate scheduler, a common recipe for Transformers \cite{grattafiori2024llama3herdmodels,oquab:hal-04376640} and other modern deep learning models \cite{10205236,10.5555/3692070.3692469}.
We set the peak learning rate to 0.0001 and the batch size to 128 in this paper.
With validation every 1000 steps, the model weights with the smallest validation loss will be adopted.

\subsection{Matching Algorithm}
Most prior studies imposed tight constraints: only one or all individuals carry sensors, matching is finalized within a predefined period, etc.
However, these assumptions are detached from real-world operations.
We develop a new matching algorithm based on insights from on-site experiments and observations.
This algorithm supports arbitrary numbers and durations of data.

\subsubsection{Assumptions}
\begin{figure}[tb]
  \centerline{\includegraphics[width=\linewidth]{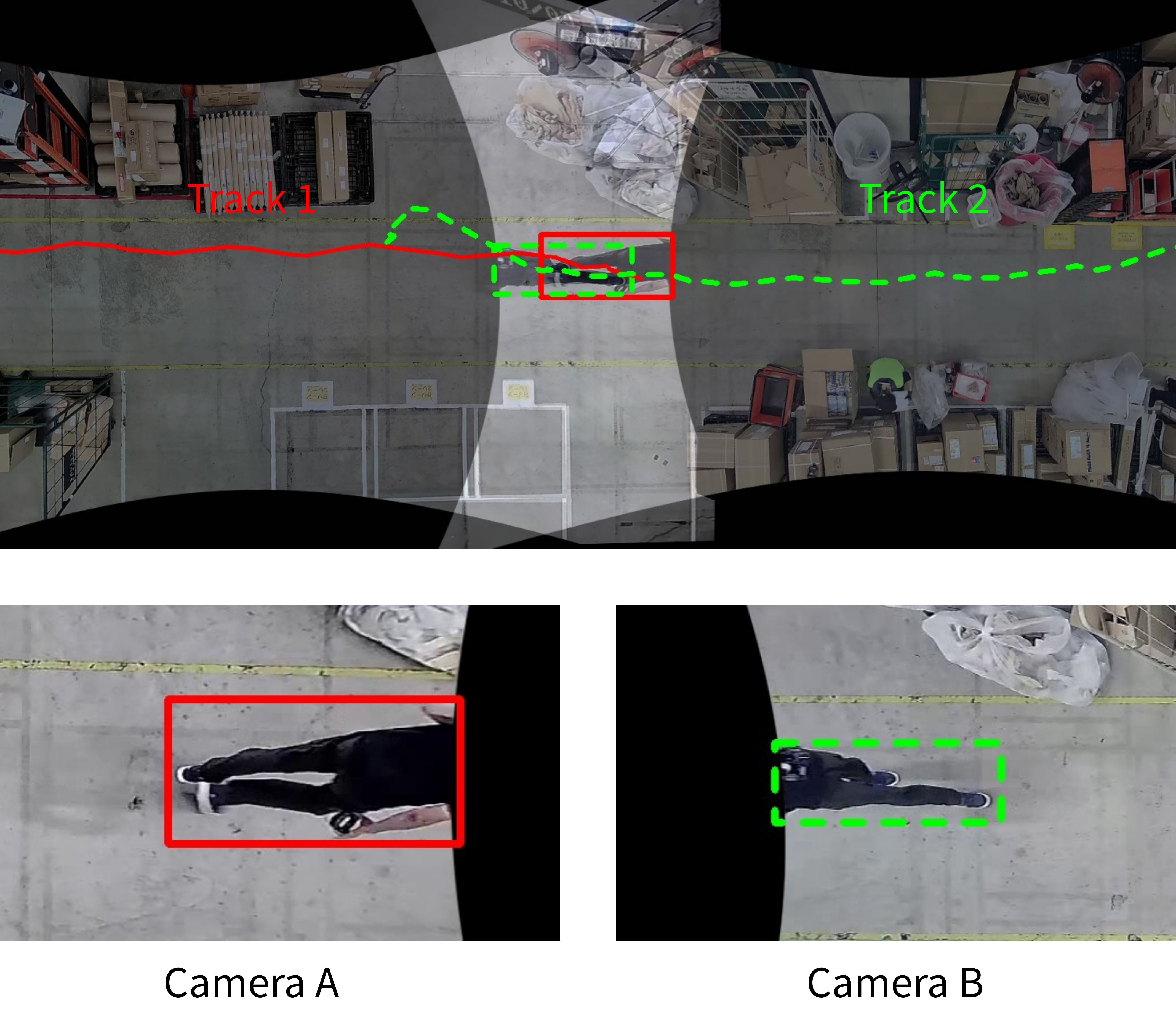}}
  \caption{Tracking duplication example.}
  \label{fig:dup}
\end{figure}

It can be unrealistic to expect everyone to carry sensors.
For instance, warehouses are open environments where external personnel such as truck drivers can enter.
Conversely, workers carrying sensors often leave camera views for work or breaks.
Visual tracking can be interrupted at occlusion as well as such out-of-view.
Moreover, tracking duplication may occur in overlapping regions among cameras, as shown in Fig. \ref{fig:dup}.
In this case, prohibiting the assignment of multiple simultaneous trajectories to the same sensor may invite more matching failures.
Based on these analyses, we define two rules below.
These permissive assumptions offer the potential for applications to various complex environments.
\begin{itemize}
  \item Every trajectory corresponds to one sensor or none.
  \item Each time point of each sensor data corresponds to any number of trajectories or none.
\end{itemize}

\subsubsection{Logic}
Our target environments involve many people compared to the prior studies.
In addition, the simultaneous presence of multiple stationary people occurs frequently.
These circumstances make it difficult to complete person identification within a limited time.
Thus, we design the matching algorithm to defer decisions for unreliable or uncertain data.
Based on the probabilities and reliabilities predicted by the model, the algorithm associates visual tracking trajectories with corresponding sensor measurements through the following procedure.

Now, for every combination of a trajectory $t$ and simultaneous sensor measurements $m$, we have the sequences of probabilities $\bm{P}^{t,m} = \left\{p^0, p^1, ...\right\}$ and reliabilities $\bm{R}^{t,m} = \left\{r^0, r^1, ...\right\}$ across the time window steps.
To begin with, time points $\bm{I}_{rel}^{t,m}$ with reliabilities higher than a threshold $R_{csdr}$ are selected.
A reliable probability average $\bar{p}_{rel}^{t,m}$ is given as the mean of probabilities over the reliable time points.
The $R_{csdr}$ determines how reliable time points will be considered.
\begin{align}
  \bm{I}_{rel}^{t,m} &= \left\{i \mid r^i > R_{csdr}\right\} \\
  \bar{p}_{rel}^{t,m} &= \frac{\Sigma_{i \in \bm{I}_{rel}^{t,m}} p^i}{\left|\bm{I}_{rel}^{t,m}\right|}
\end{align}
The reliable time points and reliable probability averages are computed for all combinations $\bm{C}$.
We can obtain $\bm{M}^t$ as the set of all measurements combined with $t$.
A trajectory $t$ will be associated with measurements $m$ if $m$ is the only sample such that the reliable probability average $\bar{p}_{rel}^{t,m}$ is higher than a threshold $P_{acpt} \geq 0.5$ among $\bm{M}^t$.
The $P_{acpt}$ determines how plausible combinations will be accepted.
% \begin{gather}
%   \bm{M}^t = \left\{m' \mid \left(t, m'\right) \in \bm{C}\right\} \\
%   \begin{split}
%     matchPositive\left(t, m\right) &:= \\
%     \left\{m' \vphantom{\bar{p}_{rel}^{t,m'}}\right. &\left.\in \bm{M}^t \mid \bar{p}_{rel}^{t,m'} > P_{acpt}\right\} = \left\{m\right\}
%   \end{split}
%   \label{eq:match}
% \end{gather}
\begin{equation}
  \bm{M}^t = \left\{m' \mid \left(t, m'\right) \in \bm{C}\right\}
\end{equation}
\begin{equation}
  \begin{split}
    matchPositive\left(t, m\right) &:= \\
    \left\{m' \vphantom{\bar{p}_{rel}^{t,m'}}\right. &\left.\in \bm{M}^t \mid \bar{p}_{rel}^{t,m'} > P_{acpt}\right\} = \left\{m\right\}
  \end{split}
  \label{eq:match}
\end{equation}
Meanwhile, a trajectory $t$ will never be associated with measurements $m$ if the reliable probability average $\bar{p}_{rel}^{t,m}$ is lower than a threshold $1 - P_{acpt}$.
\begin{equation}
  matchNegative\left(t, m\right) := \ \bar{p}_{rel}^{t,m} < 1 - P_{acpt}
\end{equation}
A trajectory $t$ will be assigned to null if all combinations of $t$ are negative.
The label null indicates people not carrying sensors.
\begin{multline}
  matchNull\left(t\right) := \\
  \left\{m' \in \bm{M}^t \mid matchNegative(t, m')\right\} = \bm{M}^t
\end{multline}
Occasionally, multiple warehouse workers move together in a coordinated manner.
We intend to defer distinguishing such uncertain combinations via the uniqueness check at \eqref{eq:match}.
Although this algorithm does not ensure matching completion in a single trial within a fixed period, it aims to identify individuals over time by confirming positive and negative pairs incrementally.

\section{Dataset Creation}
This section introduces our dataset for developing and evaluating systems that identify visually tracked people wearing sensors.
The data was collected in collaboration with our partner company.
Participants provided their written consent voluntarily after receiving a full explanation of the purpose and data usage.
In this section, we will describe the dataset creation processes and also analyze its characteristics.
Key extensions regarding the dataset from our preliminary work \cite{11213222} include the following.
\begin{itemize}
  \item The test data has been expanded by a factor of 3.
  \item The dataset will be made publicly available.
\end{itemize}

\subsection{Scenario}
\begin{figure}[tb]
  \centerline{\includegraphics[width=\linewidth]{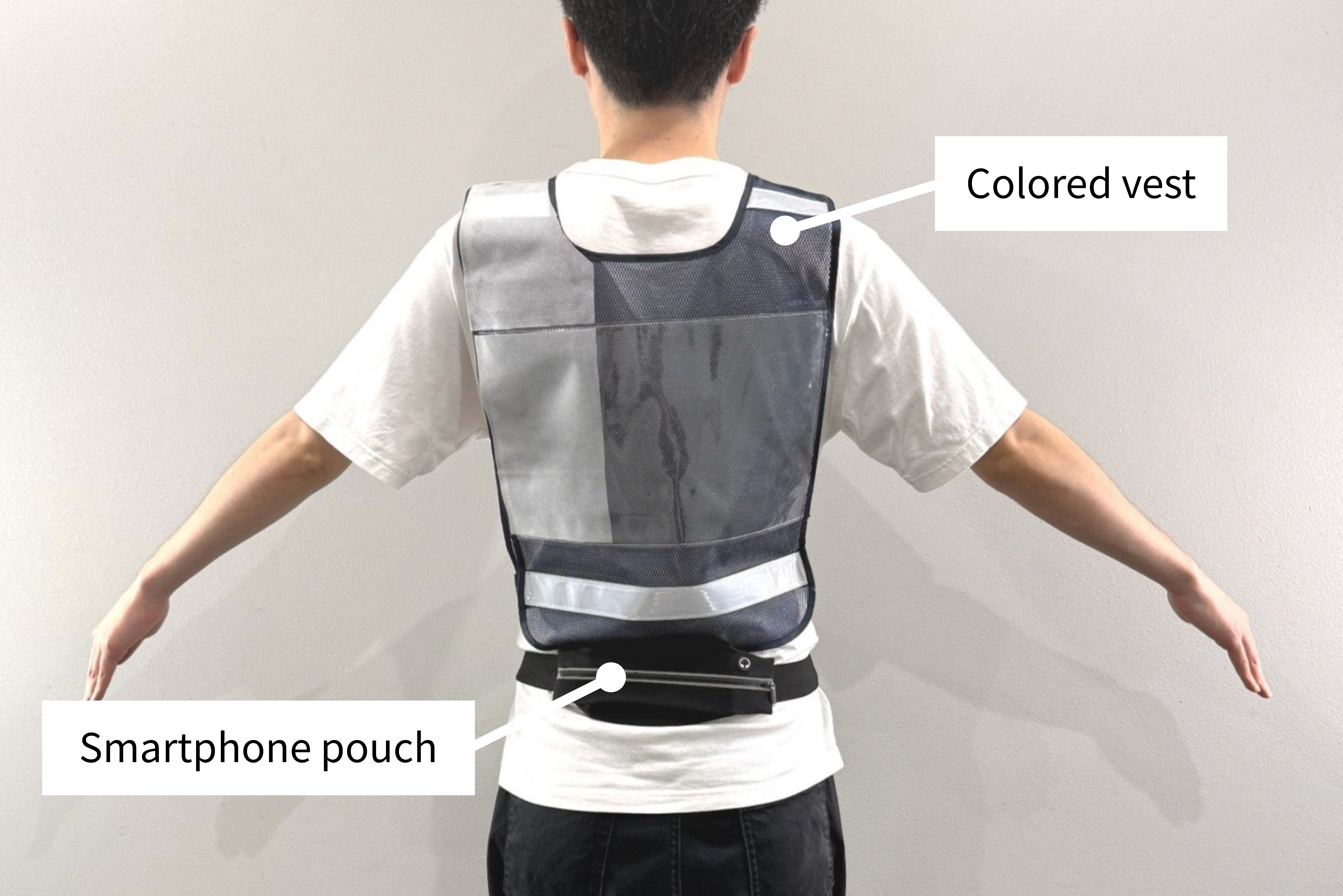}}
  \caption{Colored vest and smartphone pouch.}
  \label{fig:vest-and-pouch}
\end{figure}

Many datasets containing pedestrian trajectories and wearable sensor measurements have been created, particularly in PDR \cite{10.1145/2968219.2968276,chen2018oxioddatasetdeepinertial,8911823} and Visual--Inertial Odometry (VIO) studies \cite{8593419,11247240,GUO2026105019}.
However, most datasets focus on controlled scenarios that differ distinctly from our targets.
Thus, we developed a dataset comprising labeled visual tracking trajectories and sensor measurements in a logistics warehouse.
29 workers assigned to the inbound area performed their tasks while wearing vests and pouches with smartphones inside, as shown in Fig. \ref{fig:vest-and-pouch}.
Each vest had a unique color pattern, which served as a reference for the trajectory labeling.
The dataset includes actual activities such as inspection, transportation, and sorting.
In addition, workers vary in body type, age, and role.
Note that not all participants (i.e., workers carrying smartphones) were present simultaneously due to the shift schedule differences.
Conversely, the cameras also captured non-participants frequently.

\subsection{Sensor Measurements from Smartphones}
We collected sensor measurement data with smartphones (ASUS Zenfone 8, Android 13), including acceleration, gravitational acceleration, and angular velocity.
These data are available via Android Sensor Framework API \footnote{\url{https://developer.android.com/develop/sensors-and-location/sensors/sensors_overview}}.
The smartphones were attached to the workers' lower backs in landscape orientations, not interfering with their operations.
We got them to enter their IDs before the measurement to associate the data with the worker identities.

\subsection{Visual Tracking Trajectories from Fixed Cameras}
We also gathered video footage from wide-angle RGB cameras (H.View HV-800G2A5 \footnote{\url{https://hviewsmart.com/products/h-view-colorcam-4k-bullet-ai-camera-with-color-night-vision-hv-800g2a5}}) mounted vertically downward on the ceiling.
In this paper, we employed 19 cameras covering the inbound area of approximately 29 \texttimes \ 18 m\textsuperscript{2}.
% We recorded the videos in full HD resolution at 5 fps and up to 8 Mbps.
They synchronized every hour and streamed video in full HD resolution at 5 fps and up to 8 Mbps.
We applied Optical Character Recognition (OCR) to extract the overlaid timestamps and corrected temporal misalignments in the recordings caused by frequent frame drops.

First of all, we undistorted the videos with Double Sphere camera models \cite{8491007}.
% Then, we predicted worker bounding boxes with an extra-large-sized YOLOv8 detection model \cite{yolov8_ultralytics}.
Then, we predicted worker bounding boxes with a YOLOv8 detection model \cite{yolov8_ultralytics}.
The model weights had been previously tuned using both manually annotated and semi-automatically synthesized data \cite{10494498,10.1145/3675094.3678447}.
Subsequently, we projected the bounding boxes onto the world coordinate system and performed multi-camera tracking \cite{mori2025multicameraworkertrackinglogistics} customized from ByteTrack \cite{10.1007/978-3-031-20047-2_1}.
Afterward, we subsampled the frames by skipping every other frame to alleviate the succeeding correction effort, resulting in a trajectory frequency of 2.5 Hz.
Lastly, we manually fixed the tracking failures except for fragmentation due to out-of-view and set ID labels to the trajectories by referring to the vest patterns.
We conducted the labeling on trajectories over 1 hour during a peak period involving the most workers.

\subsection{Data Analysis}
With the labeled data, we allocated the first 30 minutes for model training and parameter selection (tune data) and the remaining 30 minutes for testing (test data).
The tune and test data contain approximately 14 and 13 hours of sensor measurements, respectively.
These durations include when out of view.
In fact, one participant never appeared in the video throughout the test data period while being measured by the sensor.
Tables \ref{tab:tune-data} and \ref{tab:test-data} summarize key statistics of visual tracking trajectories for each data.
The whole dataset contains more than 25 hours and 38 km of trajectories from participants or non-participants in total.

\begin{figure*}[tb]
  \centerline{\includegraphics[width=\linewidth]{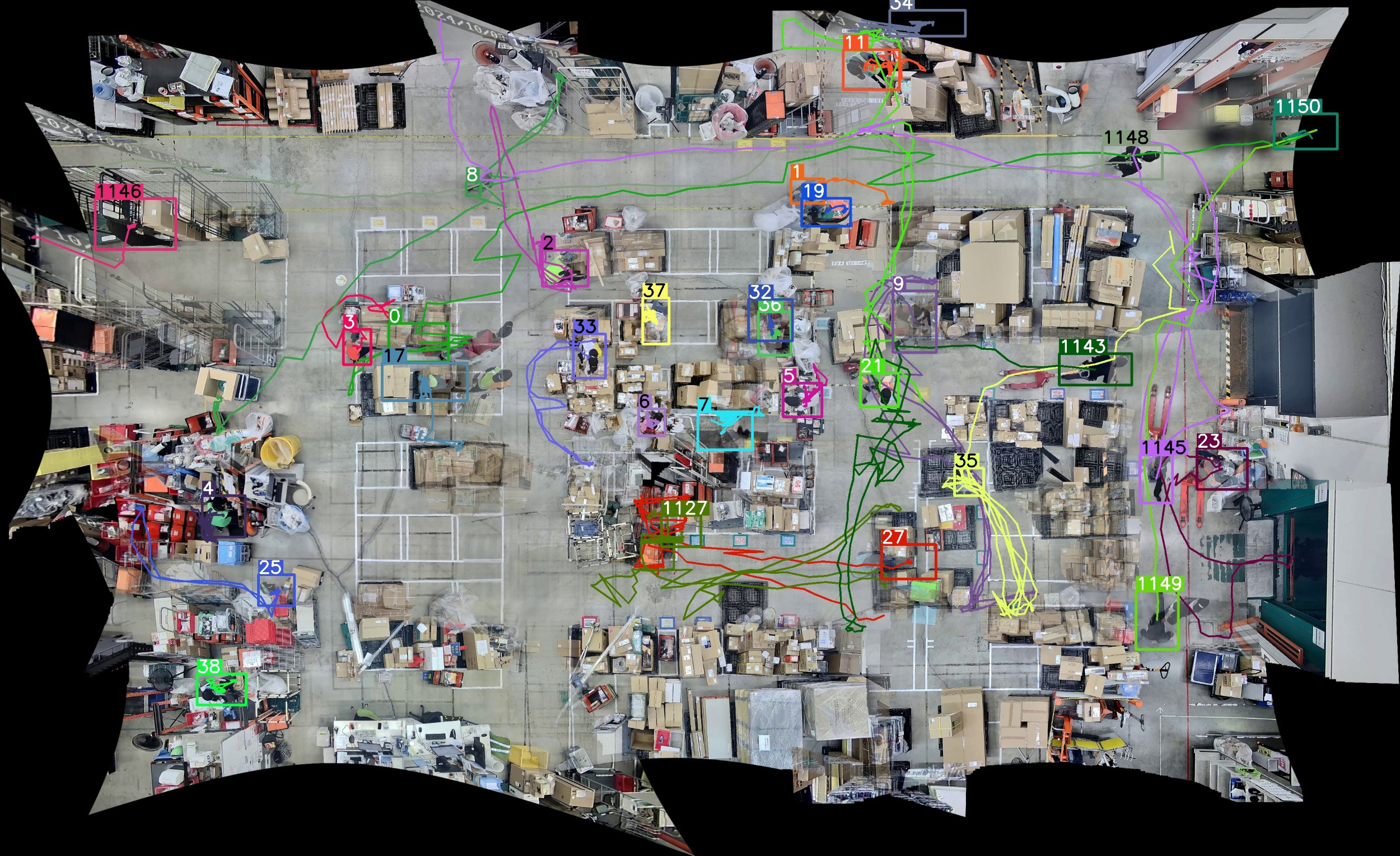}}
  \caption{Label example with last 1-minute trajectories in test data.}
  \label{fig:label}
\end{figure*}

\begin{table}[tb]
  \caption{Key Statistics of Trajectories on Tune Data}
  \begin{center}
    \begin{tabular}{|l|c|}
      \hline
      Video Duration (sec) & $1800$ \\
      \# of Unique Participants in Video & $28$ \\
      \hline
      \# of Participant Trajs & $103$ \\
      Total Time of Participant Trajs (sec) & $39057$ \\
      Quantile Times of Participant Trajs (sec) & $30$ / $122$ / $622$ \\
      Total Dist of Participant Trajs (m) & $14731$ \\
      Quantile Dists of Participant Trajs (m) & $21$ / $63$ / $191$ \\
      \hline
      \# of Non-participant Trajs & $141$ \\
      Total Time of Non-participant Trajs (sec) & $7408$ \\
      Quantile Times of Non-participant Trajs (sec) & $5$ / $18$ / $47$ \\
      Total Dist of Non-participant Trajs (m) & $6030$ \\
      Quantile Dists of Non-participant Trajs (m) & $3$ / $18$ / $37$ \\
      \hline
    \end{tabular}
    \label{tab:tune-data}
  \end{center}
\end{table}

\begin{table}[tb]
  \caption{Key Statistics of Trajectories on Test Data}
  \begin{center}
    \begin{tabular}{|l|c|}
      \hline
      Video Duration (sec) & $1800$ \\
      \# of Unique Participants in Video & $25$ \\
      \hline
      \# of Participant Trajs & $96$ \\
      Total Time of Participant Trajs (sec) & $37361$ \\
      Quantile Times of Participant Trajs (sec) & $26$ / $117$ / $553$ \\
      Total Dist of Participant Trajs (m) & $11539$ \\
      Quantile Dists of Participant Trajs (m) & $17$ / $55$ / $178$ \\
      \hline
      \# of Non-participant Trajs & $146$ \\
      Total Time of Non-participant Trajs (sec) & $7085$ \\
      Quantile Times of Non-participant Trajs (sec) & $11$ / $24$ / $45$ \\
      Total Dist of Non-participant Trajs (m) & $5956$ \\
      Quantile Dists of Non-participant Trajs (m) & $6$ / $28$ / $37$ \\
      \hline
    \end{tabular}
    \label{tab:test-data}
  \end{center}
\end{table}

Focusing on the test data, a quarter of the participant trajectories had a travel distance of shorter than 20 meters, with some under 5 meters.
The actual distances could be even shorter, as these distances were computed from raw trajectories before the smoothing.
It suggests that the data includes participants who stayed stationary most of the time.
Fig. \ref{fig:label} displays labels with the last 1-minute trajectories drawn on an image stitched from multi-camera frames at a certain time in the test data.
The label IDs above 1000 indicate non-participants.
We can see that some workers in an inspection zone, near the center of the image, have hardly moved throughout the period.
At the same time, some trajectories exhibit unnatural zigzag artifacts.
They presumably stem from lens distortion and coordinate mismatches at the camera boundaries.

\section{Evaluation and Discussion}
This section aims to demonstrate the effectiveness of CorVS+ and obtain insights for further advancement.
We will provide the evaluation settings and results, and clarify the method strengths and remaining challenges.
Key extensions regarding the evaluation from our preliminary work \cite{11213222} include the following.
\begin{itemize}
  \item Hyperparameter optimization of the model architecture has been performed.
  \item Comparison with Yan et al.'s method \cite{s24113680} has been added.
  \item An ablation study with eight variants has been added.
\end{itemize}

\subsection{Metrics}
A conventional accuracy rate is inappropriate where individuals without sensors account for a considerable portion.
We introduce new metrics to evaluate identification performance specifically for people carrying sensors, which is our primary interest.
We define Participant Precision (PP) as an extension of standard precision, the proportion of trajectories predicted correctly among all trajectories predicted as participants.
Participant Recall (PR) and Participant F1 score (PF) are also given by equations below, where $\hat{y}^j$ and $y^j$ represent predicted and actual ID labels for the $j$-th trajectory, and $\bm{L}_p$ represents the set of participant ID labels.
\begin{align}
  ParticipantPrecision &:= \frac{\left|\{j \mid \hat{y}^j \in \bm{L}_p \land \hat{y}^j = y^j\}\right|}{\left|\{j \mid \hat{y}^j \in \bm{L}_p\}\right|} \\
  ParticipantRecall &:= \frac{\left|\{j \mid y^j \in \bm{L}_p \land \hat{y}^j = y^j\}\right|}{\left|\{j \mid y^j \in \bm{L}_p\}\right|} \\
  ParticipantF\mathit{1} &:= \frac{2 \ PP \cdot PR}{PP + PR}
\end{align}
Here, the $\hat{y}^j$ and $y^j$ will be null for non-participants.
% The $\hat{y}^j$ can also be undefined if the trajectory is shorter than the minimum model input length or the matching is not confirmed.
The $\hat{y}^j$ can also be undefined if the matching is not confirmed.
In evaluation, we treat such unmatched trajectories as incorrect.
\begin{equation}
  \forall j, \ \hat{y}^j \in \bm{L}_p \cup \left\{\text{null}, \text{undefined}\right\} \ \land \ y^j \in \bm{L}_p \cup \left\{\text{null}\right\}
\end{equation}
Additionally, to better reflect the importance of informative trajectories, we also assess weighted versions of the metrics according to the trajectory time duration.

\subsection{Model Training and Parameter Selection}
At the outset, we randomly split the tune data into training and validation subsets with an approximate 8 : 2 ratio.
Here, the individuals were assigned exclusively to either subset to prevent overfitting.
Next, we constructed positive and negative pairs from them.
In this experiment, the negative sample ratio $\rho_{neg}$ for the training subset was varied to 4, 64, and 1024 to investigate its effect on the model performance.
At the same time, the positive samples in the training subset were quadrupled by randomly masking and shifting.
The time window length $W$ was set to 600 (i.e., 1 minute) as previously mentioned.
Then, we trained the correspondence estimation models and optimized their key architecture-related hyperparameters for each $\rho_{neg}$, as effective configurations would depend on the data amount.
Specifically, Tree-structured Parzen Estimator (TPE) \cite{10.5555/2986459.2986743}, a representative Bayesian optimization algorithm, was employed with a search space listed in Table \ref{tab:cand-hp} and a sampling fraction of a quarter.
Note that several parameter combinations were constrained, informed by prior studies \cite{end-to-end-speed-estim,9393889} or necessitated by dimensional consistency.
\begin{gather}
  2 \ ch = d_{attn} = 4 \ d_{mlp} \\
  2 \ ks_s - 1 = ks_l
\end{gather}

\begin{table}[tb]
  \caption{Hyperparameter Candidates of Model Architecture}
  \begin{center}
    \begin{tabular}{|cc|c|}
      \hline
      \multirow{5}{*}{CNN Backbone} & channel num $ch$ & $16$ / $24$ / $32$ \\
      & small kernel size $ks_s$ & $3$ / $5$ \\
      & large kernel size $ks_l$ & $5$ / $9$ \\
      & stride size & $1$ \\
      & padding size & $0$ \\
      \hline
      \multirow{5}{*}{Transformer Encoder} & layer num $n_{enc}$ & $2$ / $3$ / $4$ \\
      & attention dim $d_{attn}$ & $32$ / $48$ / $64$ \\
      & attention head num $h$ & $2$ / $4$ \\
      & feed forward dim $d_{ff}$ & $64$ / $96$ / $128$ \\
      & dropout rate & $0.1$ \\
      \hline
      \multirow{2}{*}{MLP Head} & hidden dim $d_{mlp}$ & $8$ / $12$ / $16$ \\
      & dropout rate & $0.1$ \\
      \hline
    \end{tabular}
    \label{tab:cand-hp}
  \end{center}
\end{table}

\begin{figure}[t!]
  \centerline{\includegraphics[width=\linewidth]{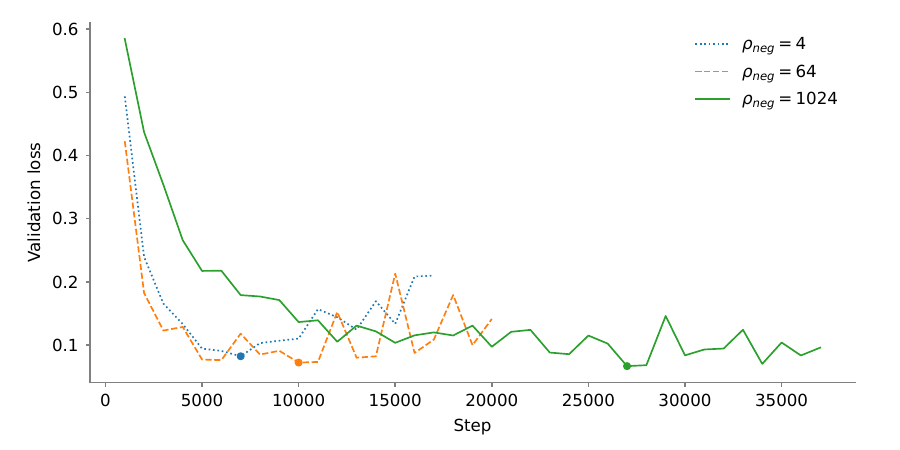}}
  \caption{Validation loss curves of optimal models.}
  \label{fig:loss}
\end{figure}

Table \ref{tab:opt-hp} presents the optimized hyperparameters and best validation losses for every $\rho_{neg}$.
Fig. \ref{fig:loss} shows the validation loss trends of those optimal models.
With larger $\rho_{neg}$, the number of learnable parameters in the optimal architecture tended to increase while the best validation loss tended to decrease.
Richer negative samples seem to have facilitated the accommodation of a broader range of data patterns.
Also, the losses showed some oscillations at $\rho_{neg} = 64, 1024$.
One possible explanation is that the predominance of negative samples led to mini-batches without any positives while amplifying the impact of each positive.
Moderate class variability across mini-batches provides regularization benefits \cite{l.2018a}.
Increasing batch sizes or capping the class ratio per batch may facilitate model training with even larger $\rho_{neg}$.

\begin{table}[tb]
  \setlength{\tabcolsep}{5.3pt}
  \caption{Optimal Hyperparameters and Best Validation Losses}
  \begin{center}
    \begin{tabular}{|c|cccccccc|c|}
      \hline
      $\rho_{neg}$ & $ch$ & $ks_s$ & $ks_l$ & $n_{enc}$ & $d_{attn}$ & $h$ & $d_{ff}$ & $d_{mlp}$ & Loss \\
      \hline
      $4$ & $16$ & $5$ & $9$ & $4$ & $32$ & $4$ & $128$ & $8$ & $0.082$ \\
      $64$ & $32$ & $5$ & $9$ & $2$ & $64$ & $4$ & $128$ & $16$ & $0.072$ \\
      $1024$ & $24$ & $5$ & $9$ & $4$ & $48$ & $4$ & $128$ & $12$ & $\bm{0.066}$ \\
      \hline
    \end{tabular}
    \label{tab:opt-hp}
  \end{center}
\end{table}

Afterward, we performed person identification on the tune data and selected the matching algorithm parameters using the optimized model at $\rho_{neg} = 1024$.
More precisely, a grid search was conducted over the reliability threshold $R_{csdr}$ and the probability threshold $P_{acpt}$ by varying them in increments of 0.05.
As a result, we adopted $\left(R_{csdr}, P_{acpt}\right)$ of (0.15, 0.90) with the best PF on the tune data.

\subsection{Baseline Comparison}
\subsubsection{Setup}
For comparison, we benchmarked two baseline methods.
The first one matches the pairs based on the agreement rate between walking events detected in visual tracking and PDR, with reference to iPAC \cite{8935339}.
It classifies the tracked subjects as walking when the movement speeds calculated from their trajectories exceed a threshold.
Similarly, it classifies the sensor wearers as walking when the walking speeds predicted by PDR exceed another threshold.
Then, it associates the pairs whose classified labels agree above a predefined proportion in decreasing order of the agreement rates.
We refer to this method as walk matching for convenience.
We borrowed the pre-trained ResNet model of RoNIN \cite{9196860} for PDR speed prediction.
The method parameters were optimized using the tune data.

Aside from that, we treat Yan et al.'s method \cite{s24113680} as the second baseline.
It uses a deep learning model to predict the correspondences, similar to CorVS+.
The model takes movement velocity from visual tracking trajectories, as well as acceleration, angular velocity, and orientations from wearable sensors, and outputs their correspondence probabilities.
After updating the probabilities in light of other pairs, it associates the pairs with probabilities higher than a threshold in decreasing order.
This method uses a brief sliding window of 1 second to segment arbitrary-length data into the model inputs.
Also, it never assigns multiple trajectories to a single sensor simultaneously, assuming no tracking duplication.
We implemented the method ourselves in accordance with the paper description.
For unspecified configurations, such as data smoothing and detailed model structure, we aligned them with ours or followed standard conventions.
Note that we used 2D movement velocity for the model inputs instead of 3D, since 3D trajectories are unavailable in our environment.
The model hyperparameters and the probability threshold were optimized using the tune data.

\subsubsection{Results}
We performed person identification on the test data with our proposed method, CorVS+, and the baseline methods.
Table \ref{tab:comp-mets} shows the metric values at that time.
To begin with, we will compare with the walk matching.
Although better than naive random, the walk matching still yielded poor performance.
A possible cause is imprecise speeds derived from noisy trajectories or predicted by PDR.
In contrast, CorVS+ bypassed this problem by estimating the correspondences directly from various features rather than via PDR.
Furthermore, imperfect time synchronization between cameras and smartphones may also have reduced the agreement rate in the walk matching.
On the other hand, CorVS+ seems to have absorbed such temporal misalignments through its tolerant inference by the correspondence estimation model.

Next, we will consider the differences from Yan et al.'s method \cite{s24113680}.
CorVS+ outperformed it overall, improving the PF by approximately 30\% and the weighted PF by approximately 8\%.
In particular, the improvement in the PP was notable.
To accommodate fragmented short trajectories, Yan et al.'s method relies on model inputs of a fixed brief length, whereas CorVS+ takes a variable-length approach.
Leveraging long-term information when available presumably contributed to the precise estimation of the correspondence probabilities.
In fact, the loss values (i.e., the gap between estimated probabilities and ground truth) during model training were also much lower with CorVS+, despite both methods using the same loss function.
Specifically, the best validation loss with the correspondence estimation model of CorVS+ was 0.066, as presented in Table \ref{tab:opt-hp}.
By contrast, Yan et al.'s method had 0.662, slightly better than the completely uncertain case of $-\ln 0.5 \approx 0.693$.
It manifests the fundamental difficulty of distinguishing individuals based on data snippets, especially in noisy environments.

\begin{table}[tb]
  \setlength{\tabcolsep}{4.9pt}
  \caption{Metric Values for Every Method on Test Data}
  \begin{center}
    \begin{tabular}{|c|ccc|ccc|}
      \hline
      & \multicolumn{3}{c|}{Normal} & \multicolumn{3}{c|}{Time Weighted} \\
      & PP & PR & PF & PP & PR & PF \\
      \hline
      Walk matching & $0.091$ & $0.229$ & $0.131$ & $0.220$ & $0.262$ & $0.239$ \\
      Yan et al. \cite{s24113680} & $0.550$ & $0.625$ & $0.585$ & $0.861$ & $\bm{0.971}$ & $0.913$ \\
      CorVS+ & $\bm{0.795}$ & $\bm{0.729}$ & $\bm{0.761}$ & $\bm{0.995}$ & $0.970$ & $\bm{0.982}$ \\
      \hline
    \end{tabular}
    \label{tab:comp-mets}
  \end{center}
\end{table}

CorVS+ resulted in a narrowly worse score than Yan et al.'s method only in the weighted PR.
Yet we tuned the matching algorithm parameters to maximize the PF in this experiment.
This performance balance is controllable to some extent by emphasizing specific metrics during parameter tuning.
In addition, Yan et al. do not account for tracking duplication.
This assumption may have favored their reported results, as this experiment used the corrected trajectory data to assess identification performance independently of tracking systems.
However, tracking duplication around camera boundaries can be a substantial factor in practice.
Given this, the superiority of CorVS+ may become further pronounced when applied to raw trajectory data.
The contributions of other differences, including the incorporation of the activity-based reliability, will be examined in the next subsection.

\subsection{Ablation}
\subsubsection{Setup}
To understand the roles of individual components in CorVS+, we carried out an ablation study.
We defined eight additional variants below in three terms: reliability awareness, data augmentation strategy, and model input length.
The first one negates the reliability consideration; in other words, it fixes the reliability threshold $R_{csdr}$ to 0 during matching.
This variant inherits the model weights from the full version (i.e., the original CorVS+) to ensure fair comparison.
The next two turn off either random masking or shifting when augmenting positive samples.
These variants maintain the same sample count and train new models with the same architecture as the full version.
The subsequent three restrict input lengths to constant rather than variable, specifically 100, 300, and 600 (i.e., 10, 30, and 60 seconds).
These variants train new models with the same architecture as the full version.
Yet the eligible sample quantity decreases as the model input length gets longer.
The last two switch between those fixed-length models depending on each sample length.
More precisely, they selectively use the model with the longest input length that does not exceed the given sample length.
The former reuses the matching algorithm parameters optimized individually for each input length, whereas the latter adopts holistic optima across three input lengths.
\begin{enumerate}
  \item No reliability consideration
  \item No random masking in positive sample augmentation
  \item No random shifting in positive sample augmentation
  \item Fixed length of 100
  \item Fixed length of 300
  \item Fixed length of 600
  \item Multiple lengths with isolated matching parameters
  \item Multiple lengths with unified matching parameters
\end{enumerate}
In this experiment, these variants shared the negative sample ratio $\rho_{neg}$ of 1024.
All other conditions, such as the data splitting and model training configurations, are also consistent with the full version unless otherwise noted.
Table \ref{tab:match-param} reports the matching algorithm parameters adjusted on the tune data.

\begin{table}[tb]
  \caption{Optimal Matching Algorithm Parameters for Every Variant on Tune Data}
  \begin{center}
    \begin{tabular}{|l|cc|}
      \hline
      & $R_{csdr}$ & $P_{acpt}$ \\
      \hline
      No reliability & -- & $0.90$ \\
      No masking & $0.05$ & $0.85$ \\
      No shifting & $0.15$ & $0.95$ \\
      Fixed 100 & $0.05$ & $0.80$ \\
      Fixed 300 & $0.35$ & $0.80$ \\
      Fixed 600 & $0.10$ & $0.75$ \\
      Multiple isolated & $0.05$ / $0.35$ / $0.10$ & $0.80$ / $0.80$ / $0.75$ \\
      Multiple unified & $0.05$ & $0.80$ \\
      Full & $0.15$ & $0.90$ \\ 
      \hline
    \end{tabular}
    \label{tab:match-param}
  \end{center}
\end{table}

\subsubsection{Results}
We performed person identification on the test data with every variant.
Table \ref{tab:abl-mets} shows the metric values at that time.
The point of departure is the activity-based reliability.
The full version achieved higher scores than the no-reliability variant across all metrics.
In this experiment, both used the same model weights, and the optimal probability threshold $P_{acpt}$ also coincided, according to Table \ref{tab:match-param}.
Hence, the performance distinction is entirely attributable to their only difference, the reliability consideration.
The no-reliability variant exhibited a particular decline in recall.
Here, the inclusion of unreliable probabilities from stationary periods would dilute informative signatures from active periods through the probability aggregation in the matching process.
This weak discriminativeness likely made it difficult to narrow down the candidate pairs, hindering completion of matching.

Next, we turn to a consideration of random masking and shifting in positive sample augmentation.
The full version demonstrated the superior results in all metrics.
Model training with a wide variety of positive samples appears to have enhanced the overall performance.
Since we weighted loss values by the inverse of class distribution, the larger the negative sample ratio $\rho_{neg}$, the greater the relative influence of each positive sample on model update.
Enriching positive sample patterns could be particularly beneficial for generalization under large $\rho_{neg}$.

\begin{table}[tb]
  \setlength{\tabcolsep}{4.4pt}
  \caption{Metric Values for Every Variant on Test Data}
  \begin{center}
    \begin{tabular}{|l|ccc|ccc|}
      \hline
      & \multicolumn{3}{c|}{Normal} & \multicolumn{3}{c|}{Time Weighted} \\
      & PP & PR & PF & PP & PR & PF \\
      \hline
      No reliability & $0.791$ & $0.708$ & $0.747$ & $0.994$ & $0.910$ & $0.950$ \\
      No masking & $0.713$ & $0.646$ & $0.678$ & $0.979$ & $0.869$ & $0.921$ \\
      No shifting & $0.756$ & $0.615$ & $0.678$ & $0.987$ & $0.880$ & $0.931$ \\
      Fixed 100 & $0.800$ & $0.625$ & $0.702$ & $0.990$ & $0.796$ & $0.883$ \\
      Fixed 300 & $0.935$ & $0.604$ & $0.734$ & $\bm{0.995}$ & $0.964$ & $0.979$ \\
      Fixed 600 & $\bm{0.946}$ & $0.552$ & $0.697$ & $0.991$ & $0.954$ & $0.972$ \\
      Multiple isolated & $0.761$ & $0.698$ & $0.728$ & $0.979$ & $0.964$ & $0.972$ \\
      Multiple unified & $0.782$ & $0.708$ & $0.743$ & $0.983$ & $0.968$ & $0.975$ \\
      Full & $0.795$ & $\bm{0.729}$ & $\bm{0.761}$ & $0.995$ & $\bm{0.970}$ & $\bm{0.982}$ \\
      \hline
    \end{tabular}
    \label{tab:abl-mets}
  \end{center}
\end{table}

Then, we will examine the effect of model input length.
The fixed-length variants tended to yield higher PP but lower PR.
Here, the fixed-length models cannot handle samples shorter than their own lengths.
From Table \ref{tab:test-data}, a quarter of participant trajectories are 26 seconds or shorter.
It implies the fixed-length variants of 300 and 600 (i.e., 30 and 60 seconds) would not achieve a PR above 0.75, even if they associated all eligible samples correctly.
On the flip side, they target only longer samples, which are relatively easy to estimate the correspondences.
This restricted scope likely led to the high PP at the expense of the PR.
On the other hand, the full version struck a reasonable balance between the PP and PR by variable-length processing that supports arbitrary-length samples.

Looking at the differences among the fixed-length variants, longer models resulted in higher PP and lower PR.
The aforementioned effect of the limited eligible samples appears to have strengthened as the model length increased.
Notably, the 600-length variant was the best in the PP and at the same time the worst in the PR among all variants.
As a result, the intermediate length of 300 brought the highest PF among the fixed-length variants.
Regarding the time-weighted metrics, the 100-length variant showed only limited gains over the normal metrics.
The time weighting accentuates the performance on longer samples.
The short model length of 100 was likely insufficient to leverage their long-term information, thereby failing to effectively benefit from the accentuation.

We now focus on the multiple-length variants.
Switching between multiple fixed-length models behaved more like a variable-length model, moderating the skew between the PP and PR.
At first glance, it might seem counterintuitive that doing so resulted in lower PP and higher PR than any of the individual models; however, it is actually plausible.
For instance, the PP of 0.800 for the 100-length variant reflects its performance on all samples longer than 10 seconds.
This value drops to 0.391 when confined to samples shorter than 30 seconds.
Conversely, the PR would improve by applying the longest possible model to each sample rather than the shortest to all samples.
In the end, the multiple-length variant with unified matching algorithm parameters achieved a higher PF than every fixed-length variant, yet the results remained modest.
The inconsistency in predicted probability distributions across the individual models may have undermined the synergy.
In contrast, the full version can handle all samples uniformly, including those shorter than 10 seconds.
Furthermore, it can better leverage contexts of various lengths.
For example, given a sample length of 299, the multiple-length variants slice it into 100, whereas the full version accepts it as is.

\subsection{Limitations and Opportunities}
In the following discussions, we will explore further performance enhancement of CorVS+.
From Table \ref{tab:abl-mets}, the full version marked the best scores in the PR, PF, weighted PR, and weighted PF.
Notably, all weighted metrics approached 1.
This result suggests that most errors arose from relatively short trajectories.
In other words, there is still room for improvement in such cases.
Specifically, the PR remained low compared to the PP.
Table \ref{tab:conf-mat} is the confusion matrix of the full version on the test data.
It indicates that the predominant factors in the PR degradation were unmatched trajectories, rather than those associated with incorrect sensors or misclassified as non-participants.
The algorithm could not determine a unique sensor to associate with.
Here, integrating extra modalities may enrich clues for matching.
One promising approach is to utilize existing Wi-Fi signals \cite{9763783}.
As previously noted, it is difficult to estimate locations robustly with trilateration or fingerprinting in warehouse inbound areas.
Nevertheless, within proximity to the access points, strong signals are available with relative stability.
RSS information from sensors may help narrow down the candidate pairs in some situations.

\begin{table}[tb]
  \setlength{\tabcolsep}{3.8pt}
  \caption{Confusion Matrix on Test Data}
  \begin{center}
    \begin{tabular}{|cc|cccc|}
      \hline
      & & \multicolumn{4}{c|}{Predicted} \\
      & & \multicolumn{2}{c}{Participant} & \multirow{2}{*}{Non-participant} & \multirow{2}{*}{Unmatched} \\
      & & Correct & Wrong & & \\
      \hline
      \multirow{2}{*}{Actual} & Participant & $70$ & $3$ & $0$ & $23$ \\
      & Non-participant & -- & $15$ & $63$ & $68$ \\
      \hline
    \end{tabular}
    \label{tab:conf-mat}
  \end{center}
\end{table}

Furthermore, the current way does not always choose the optimal matching algorithm parameters $R_{csdr}$ and $P_{acpt}$ for the test data.
In fact, the best $(R_{csdr}, P_{acpt})$ of the full version on the test data was (0.10, 0.90), yielding even higher PF.
It can be attributable to discrepancies in the distributions of model inference in addition to the data itself.
In this experiment, we used the same data for model training and parameter selection, as the labeled trajectory data was limited.
However, it can impede generalization performance because the matching algorithm parameters are adjusted based on data that the model has already seen in part.
The most straightforward workaround is to prepare separate data for model training and parameter selection.
Alternatively, an approach is worth exploring to refine the parameters in response to the distributional shifts.

Another promising direction for future work is to utilize sensor measurements to reinforce visual tracking after the matching.
This experiment used the corrected trajectory data for evaluation, but tracking switches and fragmentations sometimes occur in practice, especially in crowded scenes.
Although switching can be suppressed to some extent by adjusting tracking algorithm parameters, this typically comes at the cost of increased fragmentation.
To overcome this limitation, incorporating motion information from sensors may help maintain tracking when visual information is lacking.
The persistent trajectories also potentially lead to higher recall in person identification.

Worker location data linked with identities plays a crucial role in improving productivity and work quality in warehouses.
It enables high-fidelity modeling of workers' behavior based on actual operations.
Those models can be used in simulations for optimization of resource management \cite{11342838}, proactive assessment of robot deployment \cite{9575499}, etc.
In addition, combining the identity-tagged location data with others can unlock even greater synergies.
It has been nearly impossible to identify packaged items solely based on visual appearance, particularly in high-mix warehouses that process millions of unique items.
However, inspected items in videos can be identified by cross-referencing the inspecting worker locations with their operational logs from warehouse management systems.
It boosts item tracability and would help address the package congestion issue, a chronic bottleneck in warehouse workflows.

\section{Conclusion}
In this study, we proposed CorVS+ for identity-aware localization using fixed cameras and wearable sensors.
This study stands out for its focus on challenging real-world scenarios and its incorporation of insights from on-site studies.
We presented a deep learning model that estimates the correspondence probabilities and activity-based reliabilities from arbitrary-length data.
It was accompanied by training techniques for effective inter-modal learning.
We also presented a matching algorithm that incrementally confirms positive and negative pairs.
It accommodates practical situations such as the presence of external people and simultaneous similar movements.

Furthermore, we developed a dataset that reflects actual warehouse operations and challenging settings not seen in prior studies.
The baseline comparison demonstrated the overall superiority of CorVS+, elevating the PF by approximately 30\% or more.
The ablation confirmed the effectiveness of the reliability consideration, positive sample augmentation, and variable-length modeling.
At the end, the discussion pointed out the remaining rooms and promising avenues for further advancement.
This study paved the way for person identification based on visual and inertial data under industrial-scale settings.
We believe the takeaways from this study can help with digital transformation across various complex environments beyond warehouses.

\section*{Acknowledgment}
% Author contributions are as follows.
% \begin{enumerate}
%   \item A. Jones: Conceptualization; software; validation
%   \item B. Smith: Investigation; data curation; writing
% \end{enumerate}
The authors thank TRUSCO Nakayama Corporation for their financial support, provision of the environment, and cooperation in the experiments.
The results were analyzed and interpreted independently by the authors to ensure objectivity.

\section*{References}
\bibliographystyle{IEEEtran}
\bibliography{references}

\begin{IEEEbiography}[{\includegraphics[width=1in,height=1.25in,clip,keepaspectratio]{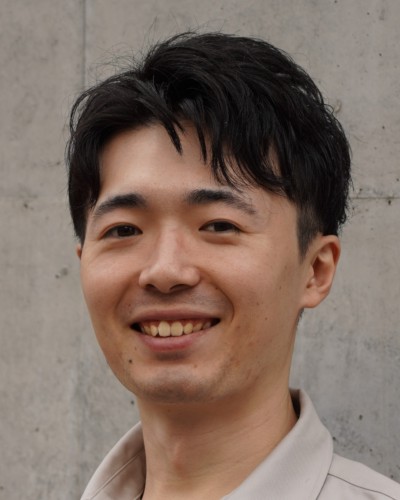}}]{Kazuma Kano}
  received B.E. degree in Electrical Engineering, Electronics, and Information Engineering and M.E. degree in Information and Communication Engineering from Nagoya University, Japan, in 2022 and 2024, respectively.
  He is currently pursuing Ph.D. degree at the same university.
  His research interests include deep learning-based indoor positioning and annotation-efficient visual recognition.
\end{IEEEbiography}

\vfill

\begin{IEEEbiography}[{\includegraphics[width=1in,height=1.25in,clip,keepaspectratio]{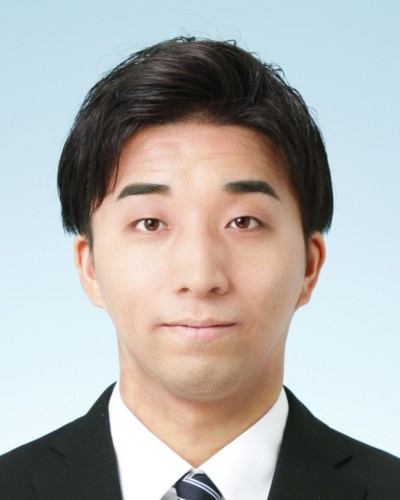}}]{Yuki Mori}
  received his B.E. degree from Nagoya University, Japan, in 2024.
  From 2024, he has been a master's student at the Graduate School of Engineering, Nagoya University.
  His research interests include ubiquitous computing and computer vision.
\end{IEEEbiography}

\begin{IEEEbiography}[{\includegraphics[width=1in,height=1.25in,clip,keepaspectratio]{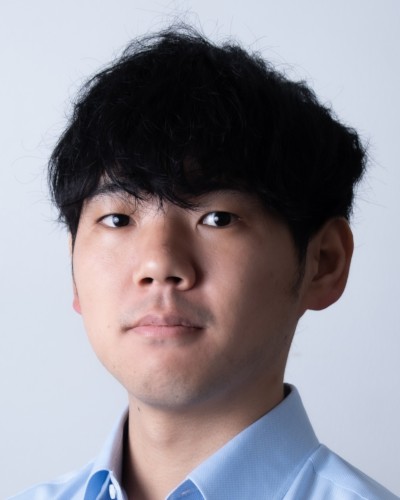}}]{Shin Katayama}
  received Ph.D. degree in engineering from Nagoya University, Japan, in 2023.
  His current research interests include human--computer interaction, ubiquitous computing systems, and affective computing.
\end{IEEEbiography}

\begin{IEEEbiography}[{\includegraphics[width=1in,height=1.25in,clip,keepaspectratio]{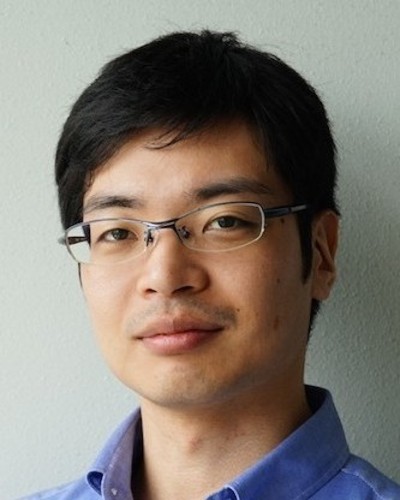}}]{Kenta Urano} (Member, IEEE)
  received his B.E., M.E., and Ph.D. degrees in engineering from Nagoya University, Japan, in 2016, 2018, and 2021, respectively.
  From 2021, he is an assistant professor in Graduate School of Engineering, Nagoya University.
  His research interests include location-based system, human activity recognition, real-world data modeling, and biosignal entertainment computing.
\end{IEEEbiography}

\begin{IEEEbiography}[{\includegraphics[width=1in,height=1.25in,clip,keepaspectratio]{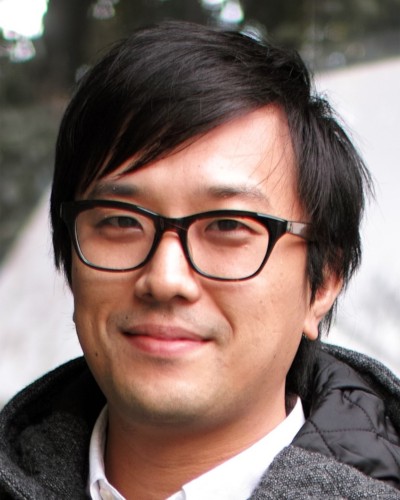}}]{Takuro Yonezawa} (Member, IEEE)
  is an associate professor in Graduate School of Engineering, Nagoya University, Japan.
  He received Ph.D. degree in the Media and Governance from Keio University in 2010.
  His research interests are the intersection of the distributed systems, human--computer interaction, and sensor/actuator technologies.
  He is a member of IPSJ, IEICE and ACM.
\end{IEEEbiography}

\begin{IEEEbiography}[{\includegraphics[width=1in,height=1.25in,clip,keepaspectratio]{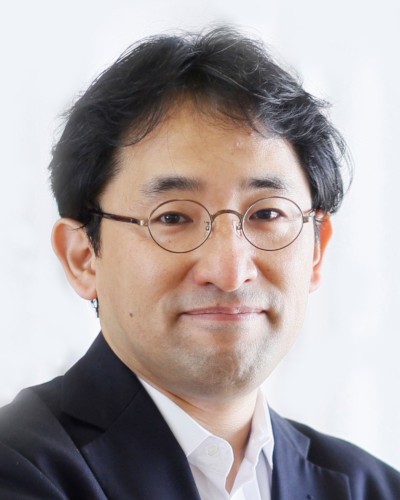}}]{Nobuo Kawaguchi} (Member, IEEE)
  received his B.E., M.E., and Ph.D. degrees in computer science from Nagoya University, Japan, in 1990, 1992, and 1995, respectively.
  From 1995, he was an associate professor in the Department of Electrical and Electronic Engineering and Information Engineering, School of Engineering, Nagoya University.
  Since 2009, he has been a professor in the Graduate School of Engineering, Nagoya University.
  His research interests are in the areas of human activity recognition, smart environmental system, and ubiquitous communication system.
\end{IEEEbiography}

\vfill

\end{document}